\documentclass[conference]{IEEEtran}
\IEEEoverridecommandlockouts
\usepackage{cite}
\usepackage{times}
\usepackage{amsmath,amssymb,amsfonts}
\usepackage{algorithmic}
\usepackage{graphicx}
\usepackage{textcomp}
\usepackage{xcolor}
\usepackage{array}
\usepackage{makecell} 
\usepackage{tabularx} 
\usepackage{booktabs} 
\usepackage{float}  
\usepackage{graphicx}  

\def\BibTeX{{\rm B\kern-.05em{\sc i\kern-.025em b}\kern-.08em
    T\kern-.1667em\lower.7ex\hbox{E}\kern-.125emX}}
\begin{document}

\title{ManchuTTS: Towards High-Quality Manchu Speech Synthesis via Flow Matching and Hierarchical Text Representation}
\author{
    \IEEEauthorblockN{
        Suhua Wang\textsuperscript{1},
        Zifan Wang\textsuperscript{2},
        Xiaoxin Sun\textsuperscript{2},
        D.J. Wang\textsuperscript{2*},
        Zhanbo Liu\textsuperscript{2},
        Xin Li\textsuperscript{3}
    }
    \IEEEauthorblockA{
        \textsuperscript{1}Department of Computer Science, Changchun Humanities and Sciences College, Changchun 130117, China \\
        \textsuperscript{2}School of Information Science and Technology, Northeast Normal University, Changchun 130117, China \\
        \textsuperscript{3}College of Optical Science and Engineering, Zhejiang University, Hangzhou 310027, China \\
        Email: yuruixin114@nenu.edu.cn
    }
}

\maketitle
\begin{abstract}
As an endangered language, Manchu presents unique challenges for speech synthesis, including severe data scarcity and strong phonological agglutination. This paper proposes ManchuTTS(Manchu Text to Speech), a novel approach tailored to Manchu's linguistic characteristics. To handle agglutination, this method designs a three-tier text representation (phoneme, syllable, prosodic) and a cross-modal hierarchical attention mechanism for multi-granular alignment. The synthesis model integrates deep convolutional networks with a flow-matching Transformer, enabling efficient, non-autoregressive generation. This method further introduce a hierarchical contrastive loss to guide structured acoustic-linguistic correspondence. To address low-resource constraints, This method construct the first Manchu TTS dataset and employ a data augmentation strategy. Experiments demonstrate that ManchuTTS attains a MOS of 4.52 using a 5.2-hour training subset derived from our full 6.24-hour annotated corpus, outperforming all baseline models by a notable margin. Ablations confirm hierarchical guidance improves agglutinative word pronunciation accuracy (AWPA) by 31\% and prosodic naturalness by 27\%.   
\end{abstract}


\section{Introduction}
\label{sec:intro}
\begin{figure}[t]
    \centering
    \includegraphics[width=0.48\linewidth, height=3cm]{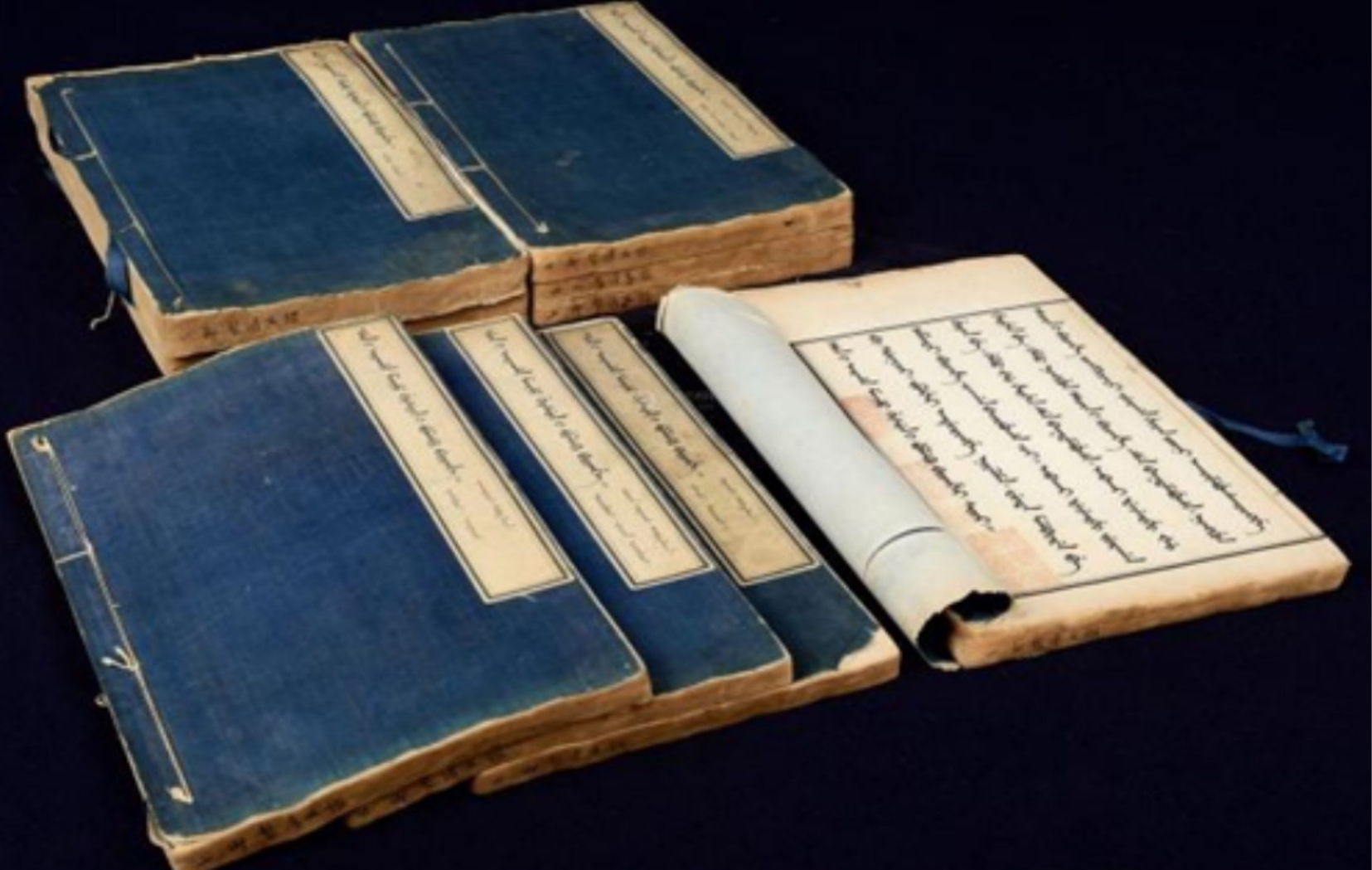}\hfill
    \includegraphics[width=0.48\linewidth, height=3cm]{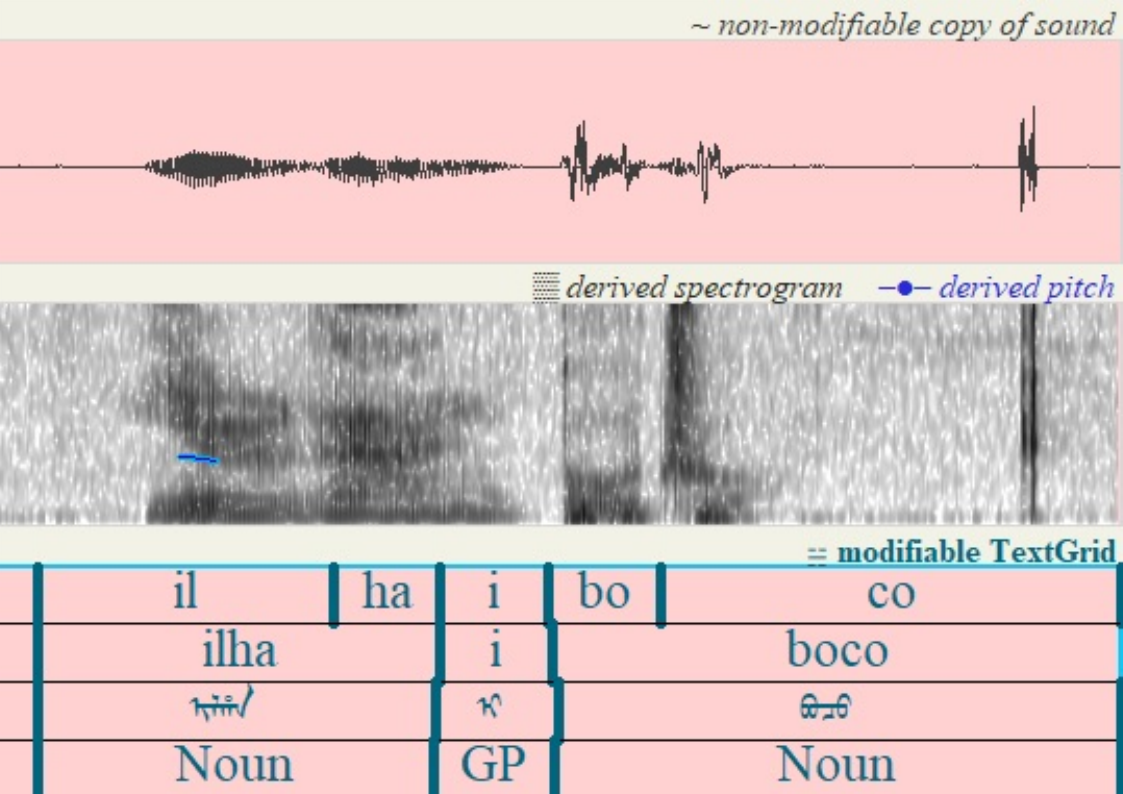}    
    \begin{minipage}{0.48\linewidth}
        \centering
        \caption{Qing Dynasty documents that have not been translated or interpreted to date.}
        \label{fig_0}
    \end{minipage}
    \hfill
    \begin{minipage}{0.48\linewidth}
        \centering
        \caption{Annotation example of Manchu speech using the Praat tool.}
        \label{fig_1}
    \end{minipage}
\end{figure}

Research on speech synthesis~\cite{1} for common languages is relatively mature, but Manchu, a UNESCO "critically endangered" language, faces severe challenges due to data scarcity and complex phonological characteristics. As the official language of China's Qing Dynasty, Manchu has left behind a vast number of historical archives awaiting translation, as shown in Fig. \ref{fig_0}. However, with fewer than 100 fluent speakers globally today, most documents remain uninterpreted, significantly hindering historical research.

Manchu speech synthesis~\cite{22} faces dual challenges from severe data scarcity and complex linguistic characteristics~\cite{21,35}. As an agglutinative language, its phonological patterns like vowel harmony are difficult for conventional TTS models to capture accurately. While methods such as LRSpeech \cite{18} offer promising directions for low-resource synthesis, they assume a minimum data threshold unavailable for truly endangered languages. Existing approaches, including flow-based models like F5-TTS \cite{3} and ReFlow-TTS \cite{28}—lack design for agglutinative morphological complexity. Technically, explicit alignment methods show high error rates (25\%–35\%) in low-resource settings, duration predictors produce rigid rhythms, and non-autoregressive architectures struggle with nuanced prosodic variations. This creates a dual challenge: overcoming both data scarcity and the alignment demands of agglutinative phonology.

\vspace{-0.5em}

To address data scarcity and agglutinative complexity, this paper propose ManchuTTS, a framework integrating hierarchical linguistic guidance with conditional flow matching \cite{7}. Unlike standard flow-based TTS, the method condition the flow vector field on three-tier linguistic features  \(c = \{c_{\text{phon}}, c_{\text{syll}}, c_{\text{pros}}\}\)
. The model employs cross-modal attention for multi-granular alignment and efficient non-autoregressive generation via conditional differential equations. Key contributions include: a hierarchical guidance mechanism for agglutinative languages; an end-to-end system based on conditional flow matching; and the first public Manchu TTS dataset with validation.
Experimental results show ManchuTTS achieves a MOS of 4.52 with only 5.2 hours training data, surpassing all baselines. Ablations validate the hierarchical design, improving agglutinative word accuracy by 31\% and prosodic naturalness by 27\%.
The main contributions of this work are as follows:
\begin{itemize}
\item This paper propose a hierarchical text feature guidance framework tailored for Manchu speech synthesis, effectively addressing its agglutinative characteristics.
\item This paper develop an end-to-end speech synthesis system based on hierarchical conditional flow matching, enabling high-quality synthesis with minimal data requirements.
\item This paper construct the first publicly available Manchu speech synthesis dataset and demonstrate the effectiveness of this approach through comprehensive experiments.
\end{itemize}

This work provides both a practical solution for Manchu speech synthesis and a methodological framework that can be extended to other endangered languages, contributing to the preservation and accessibility of linguistic heritage through technological innovation.
\vspace{-0.3em}

\section{Method}
\noindent\textbf{Theoretical Framework.}
This method establishes an implicit mapping from reference speech \(V_{\text{ref}}\) and text \(T_{\text{ref}}\) to target speech for \(T_{\text{tgt}}\), forming a ternary speech-text-acoustic relationship. To handle Manchu's agglutinative nature, the input text is decomposed into three hierarchical linguistic units. \(c_{\text{phon}}\) is phoneme-level features. \(c_{\text{syll}}\) is syllable/word-structure features. \(c_{\text{pros}}\) is prosodic features.
These are integrated into a conditional flow matching framework to guide the speech generation process.

\noindent\textbf{Conditional Flow Matching.}
This model the transformation from noise \(x_0\) to target speech \(x_1\) via a linear interpolation path:
\vspace{-0.5em}
\begin{equation}
x_t = (1 - t) \cdot x_0 + t \cdot x_1, \quad t \in [0,1].
\end{equation}
\vspace{-0.3em}
where \(x_0 \sim \mathcal{N}(0, I)\) and \(x_1\) is the target mel-spectrogram. The corresponding vector field that drives this transformation is:
\vspace{-0.3em}
\begin{equation}
u_t(x_t \mid c) = x_1 - x_0.
\end{equation}
\vspace{-0.3em}
The goal is to learn a parameterized model \(v_t(x_t \mid c; \theta)\) that approximates \(u_t\). The training objective is:
\vspace{-0.3em}
\begin{equation}
\mathcal{L}_{\text{CFM}}(\theta) = \mathbb{E}_{t, x_0, x_1, c} \left\| v_t(x_t \mid c; \theta) - (x_1 - x_0) \right\|^2.
\end{equation}
\vspace{-0.3em}
Here, \(c = \{c_{\text{phon}}, c_{\text{syll}}, c_{\text{pros}}\}\) represents the multi-level linguistic conditions. At inference, this method start from Gaussian noise and solve the ODE defined by \(v_t\) to generate speech that aligns with the input text.

\begin{figure*}[t] 
\centering
\includegraphics[width=1\linewidth]{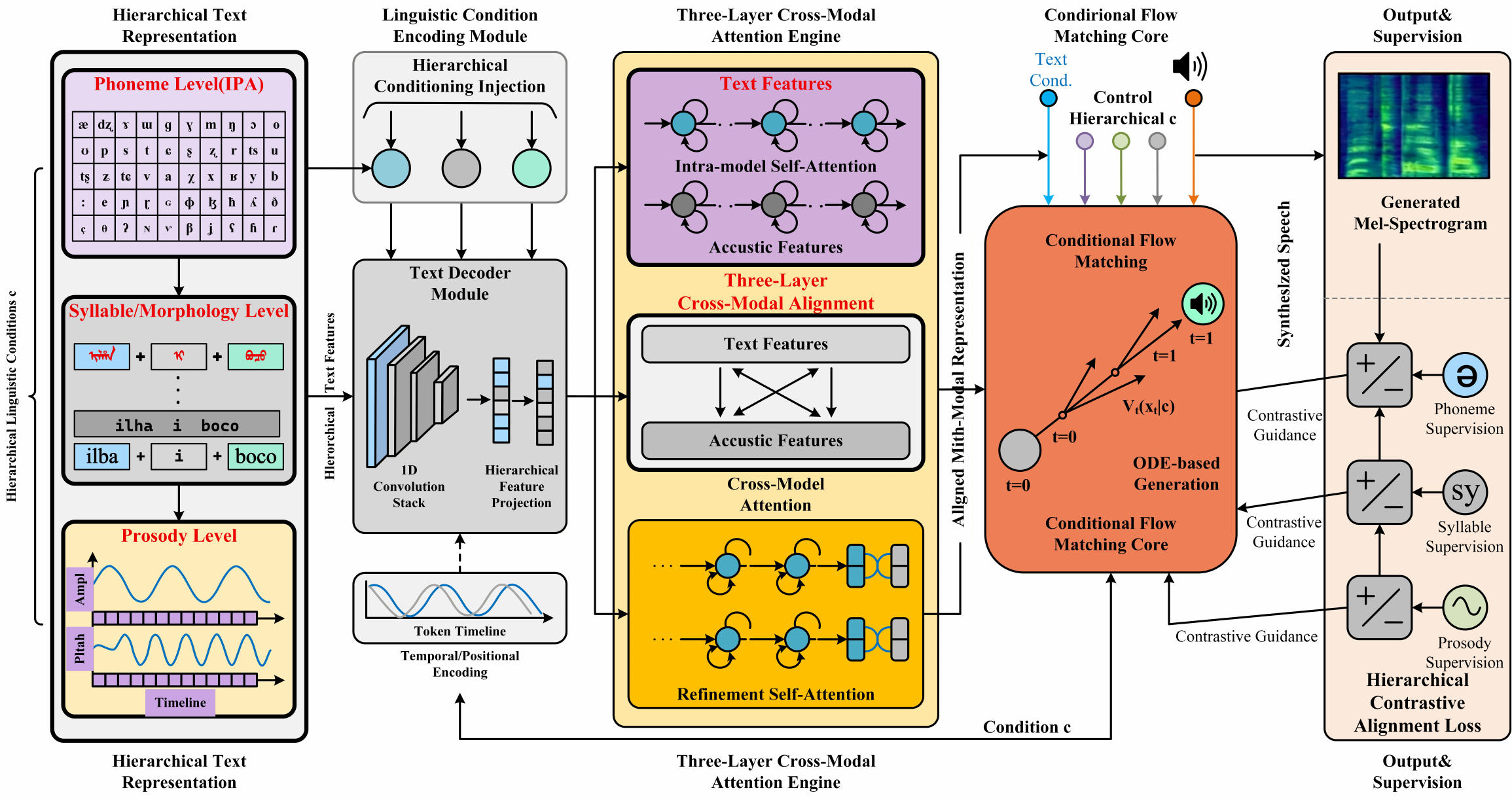}
\caption{Overall Pipeline of ManchuTTS Framework. The system encodes target text into three-level linguistic features, processes them through an 8-layer Diffusion Transformer (DiT) backbone, and generates speech via conditional flow matching.}
\label{fig_2}
\vspace{-1em}
\end{figure*}

To handle Manchu's complex agglutinative morphology, this work proposes a three-stage mechanism (phoneme-syllable-prosody). Fig.~\ref{fig_1} illustrates the multi-layer Praat annotations for an agglutinative sentence, validated by linguists. The complete synthesis pipeline of ManchuTTS is presented in Fig.~\ref{fig_2}.
\subsection{Three-Layer Cross-Modal Attention Mechanism}
This paper proposes a three-layer cross-modal attention architecture for progressive alignment between textual features \(\mathbf{F}_t\) and acoustic features \(\mathbf{F}_a\). Following a "self-attention → interaction → self-attention" paradigm, it enables stepwise cross-modal alignment.

Layer 1: Intra-modal Self-Attention
First, both textual and acoustic features undergo intra-modal self-attention to enhance their internal representations. The multi-head attention mechanism captures semantic dependencies within the textual sequence, computed as:
\begin{equation}
\mathbf{F}_t^{(1)} = \text{LayerNorm}\big(\mathbf{F}_t + \text{MHA}(\mathbf{F}_t, \mathbf{F}_t, \mathbf{F}_t)\big).
\end{equation}
Similarly, the acoustic features are processed through an analogous self-attention operation to strengthen the temporal relationships within the acoustic sequence:
\begin{equation}
\mathbf{F}_a^{(1)} = \text{LayerNorm}\big(\mathbf{F}_a + \text{MHA}(\mathbf{F}_a, \mathbf{F}_a, \mathbf{F}_a)\big).
\end{equation}
Here, 
\(\text{MHA}(Q, K, V)\) denotes the standard multi-head attention mechanism. This layer produces enhanced modality-specific features, laying the foundation for subsequent cross-modal interaction.

Layer 2: Cross-modal Cross-Attention
After obtaining internally refined features, cross-modal interaction is performed via a bidirectional cross-attention mechanism. Specifically, the textual features are updated by attending to the acoustic features as contextual information:
\begin{equation}
\mathbf{F}_t^{(2)} = \text{LayerNorm}\big(\mathbf{F}_t^{(1)} + \text{MHA}(\mathbf{F}_t^{(1)}, \mathbf{F}_a^{(1)}, \mathbf{F}_a^{(1)})\big).
\end{equation}
Symmetrically, the acoustic features are updated by attending to the textual features, ensuring mutual information exchange:
\begin{equation}
\mathbf{F}_a^{(2)} = \text{LayerNorm}\big(\mathbf{F}_a^{(1)} + \text{MHA}(\mathbf{F}_a^{(1)}, \mathbf{F}_t^{(1)}, \mathbf{F}_t^{(1)})\big).
\end{equation}
This layer serves as the core of cross-modal alignment, allowing each modality to integrate relevant information from the other and promoting feature-space fusion.

Layer 3: Intra-modal Self-Attention
Finally, to consolidate the information acquired through cross-modal interaction and further refine the internal representations, another intra-modal self-attention layer is applied. The output textual features are computed as:
\begin{equation}
\mathbf{F}_t^{\text{out}} = \text{LayerNorm}\big(\mathbf{F}_t^{(2)} + \text{MHA}(\mathbf{F}_t^{(2)}, \mathbf{F}_t^{(2)}, \mathbf{F}_t^{(2)})\big).
\end{equation}
Similarly, the output acoustic features are obtained via:
\begin{equation}
\mathbf{F}_a^{\text{out}} = \text{LayerNorm}\big(\mathbf{F}_a^{(2)} + \text{MHA}(\mathbf{F}_a^{(2)}, \mathbf{F}_a^{(2)}, \mathbf{F}_a^{(2)})\big).
\end{equation}
The resulting aligned features, \(\mathbf{F}_t^{\text{out}}\) and \(\mathbf{F}_a^{\text{out}}\), preserve the internal consistency of their respective modalities while embodying complementary cross-modal information.

\subsection{Hierarchical Text Representation}
ManchuTTS employs a three-tier text representation to address Manchu's agglutinative structure. At the phoneme level, input text is converted into IPA sequences to capture fine-grained phonetic features including vowel harmony and coarticulation patterns. The syllable/word level decomposes complex affixes into [root+suffix] structures, modeling inter-syllable transitions and root-suffix prosodic coupling to handle morphological variations. The prosody level predicts global intonation and rhythm patterns, adapting pitch contours and energy distribution according to sentence types (e.g., rising pitch for questions). Collectively, these hierarchical features c 
provide structured linguistic guidance for subsequent cross-modal alignment and conditional flow matching.
\subsection{Hierarchical Contrastive Alignment Loss}
Based on the conditional flow matching model, this paper propose a hierarchical contrastive alignment loss to enhance the consistency between the generated speech and the hierarchical linguistic conditions \(c = \{c_{\text{phon}}, c_{\text{syll}}, c_{\text{pros}}\}\). For each level of condition \(c^{(k)}\), this paper construct positive sample pairs \((x_1, c^{(k)})\) and negative sample pairs \((x_1, c^{(k)}_{\text{neg}})\), and maximize the mutual information between positive pairs. The loss function is defined as a weighted sum of the contrastive losses at each level:
\vspace{-0.3em}
\begin{multline}
\mathcal{L}_{\text{HCA}} = \sum_{k} \lambda_k \cdot \mathbb{E} \biggl[ \\
-\log \frac{\exp(s(x_1, c^{(k)}))}{\exp(s(x_1, c^{(k)})) + \sum_{c^{(k)}_{\text{neg}}} \exp(s(x_1, c^{(k)}_{\text{neg}}))} \biggr].
\end{multline}
\vspace{-0.3em}
where \(s(\cdot, \cdot)\) is a similarity function. This loss guides the generation process to better preserve fine-grained linguistic information at the phoneme, syllable, and prosodic levels.
\subsection{Model Innovations and Comparison with Existing Methods}
ManchuTTS introduces a novel "multi-granular linguistic conditional flow matching" framework. Unlike standard flow-matching TTS models (e.g., F5‑TTS, E2/E3‑TTS), This method jointly models agglutinative structures by conditioning the flow vector field on hierarchical linguistic features 
\(c = \{c_{\text{phon}}, c_{\text{syll}}, c_{\text{pros}}\}\). This coupling of linguistic priors with the underlying ODE fundamentally reshapes the probability flow path during generation. In contrast to traditional hierarchical TTS that merely uses multi‑level embeddings, this conditions directly steer both the flow‑matching loss and the hierarchical contrastive alignment loss, enabling precise implicit alignment tailored to Manchu's phonological characteristics.

\section{Experiments}
\subsection{Experimental Setup}
\noindent\textbf{Dataset.} This study employs a self-collected Manchu speech dataset comprising 6.24 hours of high-quality audio (44.1 kHz, 16-bit PCM). The data underwent rigorous quality control, including SNR filtering ($\geq 25\,\mathrm{dB}$) and expert validation. It covers 90\% of core phonemes and six intonation types. The dataset is split into training (5.2\,h, gender-balanced), validation (0.52h), and test sets (0.52h, including cross-gender and cross-complexity samples). Recordings were made with professional equipment, and all texts were verified by linguists. A subset of aligned speech-text samples will be made available for research purposes (see Table~\ref{tab:1}).

\begin{table}[t]
\vspace{-1.5em}
\caption{Dataset Partitioning}
\vspace{-0.9em}
\label{tab:1}
\begin{center}
\begin{tabular}{|l|c|c|c|}
\hline
\multicolumn{4}{|c|}{\textbf{Dataset Specifications}} \\
\hline
\textbf{Subset} & \textbf{Duration} & \textbf{Utterances} & \textbf{Description} \\
\hline
Training   & 5.2 h  & 1,892 & Model parameter training \\
\hline
Validation & 0.52 h & 236   & Hyperparameter tuning \\
\hline
Test       & 0.52 h & 236   & Final performance evaluation \\
\hline
\multicolumn{3}{|l|}{\textbf{Total}} & \textbf{2,364 utterances (6.24 h)} \\
\hline
\end{tabular}
\end{center}
\vspace{-1em}
\end{table}

\noindent\textbf{Training Configuration.} As is shown in Table~\ref {tab:2}, the model was trained for 300K steps with a global batch size of 256 on 24GB GPUs. Optimizer: AdamW with cosine decay (warmup 5K steps, peak 3e-4); gradient clipping 1.2; dropout 0.1–0.15. Architecture: 2-layer phoneme encoder, 8-layer DiT backbone (512 dim, 4 heads), 3-layer LSTM duration predictor (86.4M parameters). Inputs: 80-dim Mel spectrograms from phonemized text (vocab: 1024); speaker embedding: 64-dim; training set: gender-balanced.
\begin{table}[t]
\centering
\caption{ManchuTTS Training Hyperparameters and Hardware Configuration}
\vspace{-0.8em}
\label{tab:2}
\begin{tabular}{|l|l|}
\hline
\textbf{Configuration Item} & \textbf{Value / Description} \\
\hline
GPU & 8 × NVIDIA RTX 4090 (24 GB) \\
\hline
Global Batch Size & 256 samples $\approx$ 0.6 hours of audio \\
\hline
Total Steps & 300k \\
\hline
Optimizer & AdamW ($\beta_1=0.9$, $\beta_2=0.98$) \\
\hline
Learning Rate Schedule & \begin{tabular}{@{}l@{}}5k steps linear warm-up \\ → peak 3e-4 \\ → cosine decay to 1e-5\end{tabular} \\
\hline
Gradient Clipping & 1.2 \\
\hline
Dropout & \begin{tabular}{@{}l@{}}Attention layers 0.1 \\ Feed-forward layers 0.15\end{tabular} \\
\hline
Model Parameters & 86.4 M \\
\hline
Conv1D Phoneme Encoder & 2 layers, 256 channels \\
\hline
Diffusion Transformer & 8 layers, 4 heads, 512 hidden units \\
\hline
LSTM Duration Predictor & 3 layers, 128 hidden units \\
\hline
Vocabulary Size & 1,024 phoneme-level \\
\hline
Mel Feature Dimensions & \begin{tabular}{@{}l@{}}80-dim, 50 ms frame length, \\ 12.5 ms frame shift\end{tabular} \\
\hline
Training Duration & $\approx$ 48 hours \\
\hline
\end{tabular}
\end{table}

\noindent\textbf{Baselines.} This paper compare the method with several representative models widely used in low-resource speech synthesis and minority languages:

\begin{itemize}
\item \textbf{Tacotron 2}\cite{6}: Classic seq2seq TTS with encoder-attention decoder; data-hungry and slow in low-resource settings.

\item \textbf{FastSpeech 2}\cite{5}: Non-autoregressive model using duration/pitch prediction; faster synthesis suitable for real-time use.

\item \textbf{Glow-TTS}\cite{37}: Flow-based model with invertible networks for bidirectional text-speech conversion.

\item \textbf{VITS}\cite{38}: End-to-end TTS integrating VAE and GAN, directly generating speech from text.

\item \textbf{F5-TTS}\cite{3}: Fast, controllable model with multi-scale architecture for complex prosody.

\item \textbf{Cloning-based Voice Conversion (CBVC)}\cite{39}: Adapts pre-trained models with minimal target data via transfer learning.
\end{itemize}

\noindent\textbf{Baseline Configuration.} All models used the same data (5.2h train, 0.52h test), pronunciation dictionary, and audio parameters (44.1kHz/16-bit, 80-dim Mel) under identical low-resource settings.

Evaluation included subjective and objective methods: Mean Opinion Score (MOS) tests by 20 native speakers with ABX preference tests; objective metrics comprising Mel Cepstral Distortion (MCD), fundamental frequency Root Mean Square Error (F0-RMSE), Word Error Rate (WER), speaker similarity (SIM), and Perceptual Evaluation of Speech Quality (PESQ). All tests used the same test set with 95\% confidence intervals.

\subsection{Experimental Results and Analysis}

\subsubsection{Main Results}
\begin{figure}[t]
\centering
\includegraphics[width=0.7\linewidth]{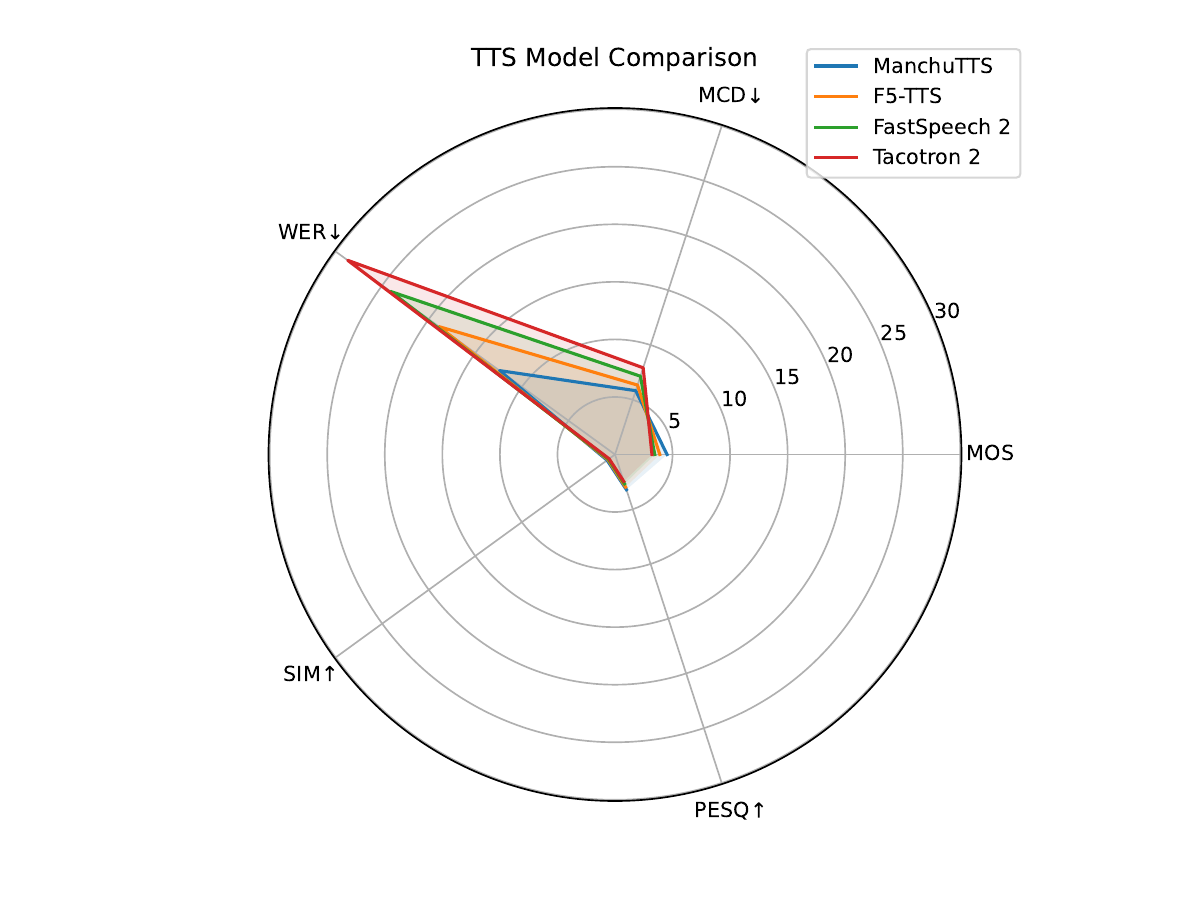}
\caption{Radar chart of multi-dimensional performance for main Manchu TTS experiments (ManchuTTS vs. baseline models).}
\vspace{-0.5em}
\label{fig_3}
\end{figure}

ManchuTTS achieves a MOS of 4.52±0.11 with only 5.2 hours training data, outperforming all baselines across objective metrics (Table~\ref{tab:3}) and showing consistent superiority (Fig.~\ref{fig_3}).

\subsubsection{Ablation Study of Hierarchical Text Feature Guidance Mechanism} 
\begin{figure}[t]
\vspace{-0.5em}
\centering
\includegraphics[width=0.75\linewidth]{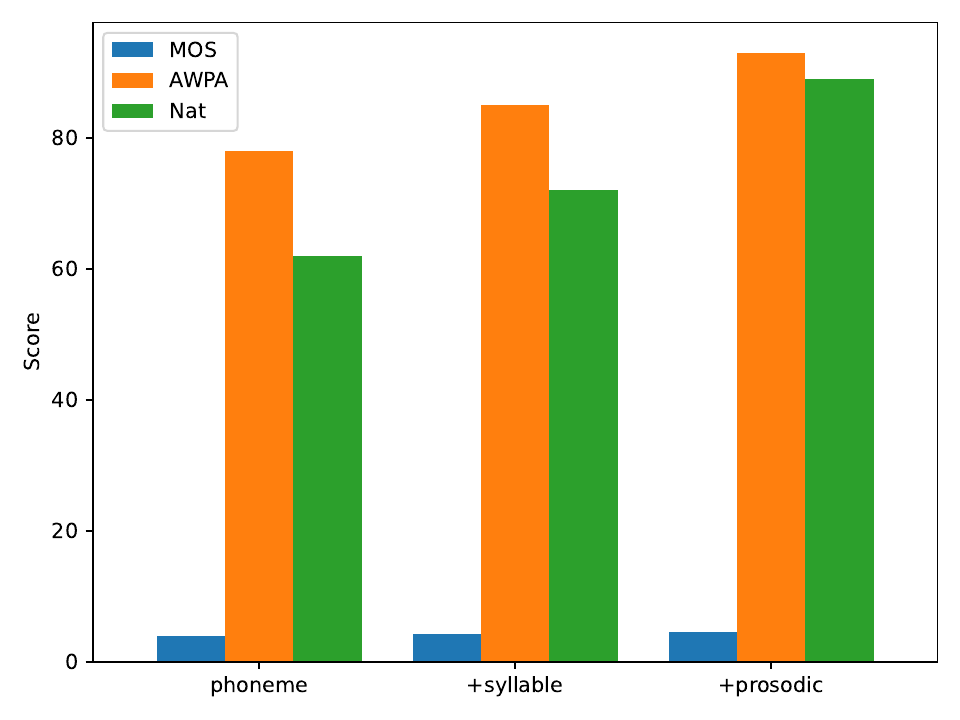}
\caption{Impact of three-level text feature guidance on MOS, AWPA, and prosodic naturalness.  
X-axis: Configuration levels ("Phoneme", "+Syllable", "+Prosody"), representing the incremental guidance conditions added step by step.  
Y-axis:  
1. MOS (Mean Opinion Score, 1–5, the higher the better);  
2. AWPA (\%, the higher the better);  
3. Prosodic naturalness (\%, the higher the better).}
\label{fig_4}
\vspace{-0.3cm}
\end{figure}

Fig.~\ref{fig_4} illustrates the performance contribution of the three-level "phoneme-syllable-prosody" guidance framework in ManchuTTS.
Using only the phoneme layer yields a MOS of 3.94. Incorporating the syllable layer increases MOS to 4.21, and enabling all three layers further improves MOS to 4.52, with notable enhancements in agglutinative word accuracy and prosodic naturalness, as summarized in Table~\ref{tab:4}.

\begin{table}[t]
\centering
\caption{Performance comparison of ManchuTTS and baseline systems (95\% CI).}
\vspace{-0.8em}
\label{tab:3}
\scriptsize
\setlength{\tabcolsep}{3pt} 
\renewcommand{\arraystretch}{1.05} 

\newlength{\modelwidth}
\setlength{\modelwidth}{1.0cm} 
\newlength{\metriccolwidth}
\setlength{\metriccolwidth}{\dimexpr(\linewidth - \modelwidth - 16\tabcolsep - 8\arrayrulewidth)/6\relax}

\begin{tabular}{|p{\modelwidth}|*{6}{>{\centering\arraybackslash}p{\metriccolwidth}|}}  
\hline
\textbf{Model} & \textbf{MOS $\uparrow$} & \textbf{MCD $\downarrow$} & \textbf{F0 $\downarrow$} & \textbf{WER $\downarrow$} & \textbf{SIM $\uparrow$} & \textbf{PESQ $\uparrow$} \\
\hline
T2 & 3.21$\pm$0.18 & 7.92$\pm$0.31 & 34.2$\pm$4.3 & 28.7$\pm$3.2 & 63.1$\pm$4.5 & 2.45$\pm$0.12 \\
\hline
FS2 & 3.45$\pm$0.15 & 7.15$\pm$0.28 & 29.8$\pm$3.7 & 24.1$\pm$2.9 & 69.8$\pm$4.1 & 2.68$\pm$0.11 \\
\hline
Glow & 3.89$\pm$0.13 & 6.34$\pm$0.25 & 22.4$\pm$2.8 & 18.9$\pm$2.4 & 75.6$\pm$3.8 & 2.97$\pm$0.10 \\
\hline
VITS & 3.95$\pm$0.12 & 6.18$\pm$0.24 & 21.7$\pm$2.7 & 17.8$\pm$2.3 & 76.9$\pm$3.7 & 3.02$\pm$0.09 \\ 
\hline
F5 & 4.12$\pm$0.11 & 5.99$\pm$0.23 & 20.3$\pm$2.5 & 15.6$\pm$2.1 & 79.4$\pm$3.5 & 3.08$\pm$0.08 \\ 
\hline
CBVC & 4.28$\pm$0.10 & 5.91$\pm$0.22 & 19.5$\pm$2.2 & 14.1$\pm$1.9 & 81.8$\pm$3.3 & 3.15$\pm$0.08 \\ 
\hline
Ours & \textbf{4.52$\pm$0.11} & \textbf{5.83$\pm$0.23} & \textbf{18.7$\pm$2.1} & \textbf{12.4$\pm$1.8} & \textbf{84.7$\pm$3.2} & \textbf{3.21$\pm$0.09} \\ 
\hline
GT & 4.68$\pm$0.09 & 4.40$\pm$0.20 & 14.2$\pm$2.0 & 11.6$\pm$0.20 & 90.5$\pm$1.5 & 3.89$\pm$0.10 \\
\hline
\end{tabular}

\vspace{0.1em}
\makebox[\linewidth][l]{
  \parbox{\linewidth}{
    \footnotesize\textit{
      Note: T2: Tacotron 2, FS2: FastSpeech 2, F5: F5-TTS, Ours: ManchuTTS, \\ 
      \ \ \ \ \phantom{Note: }GT: Human-recorded speech.
    }
  }
}
\vspace{-0.3em}
\end{table}

\begin{table}[t]
\vspace{-0.3em}
\centering
\caption{Effectiveness of Three-Level Guidance Framework}
\vspace{-0.8em}
\label{tab:4}
\scriptsize  
\begin{tabular}{|l|c|c|c|c|c|c|}
\hline
\textbf{Cfg} & \textbf{\makecell{Phoneme\\Layer}}  & \textbf{\makecell{Syllable\\Layer}}  & \textbf{\makecell{Prosody\\Layer}}  & \textbf{MOS} & \textbf{AWPA} & \textbf{PN} \\
\hline
A & \( \checkmark \) & \( \times \) & \( \times \) & 3.94$\pm$0.14 & 70.8\% & 70.4\% \\
\hline
B & \( \checkmark \) & \( \checkmark \) & \( \times \) & 4.21$\pm$0.12 & 84.7\% & 80.5\% \\
\hline
C & \( \checkmark \) & \( \checkmark \) & \( \checkmark \) & 4.52$\pm$0.11 & 92.7\% & 89.4\% \\
\hline
\end{tabular}
\vspace{0.5em}
\makebox[\linewidth][l]{
  \parbox{\linewidth}{
    \footnotesize\textit{
      Note: AWPA: Agglutinative Word Pronunciation Accuracy, PN: Prosodic \\ \phantom{Note: }Naturalness (\%).  Assessed by two native linguists via a binary cla-\\ \phantom{Note: }ssification task.
    }
  }
}
\vspace{-2em}
\end{table}

\noindent Table~\ref{tab:4} shows the impact of progressively removing guidance layers (other conditions fixed) on MOS, agglutinative word accuracy, and prosodic naturalness.

\begin{figure*}[t]
\vspace{-0.5em}
\centering
\includegraphics[width=1\linewidth]{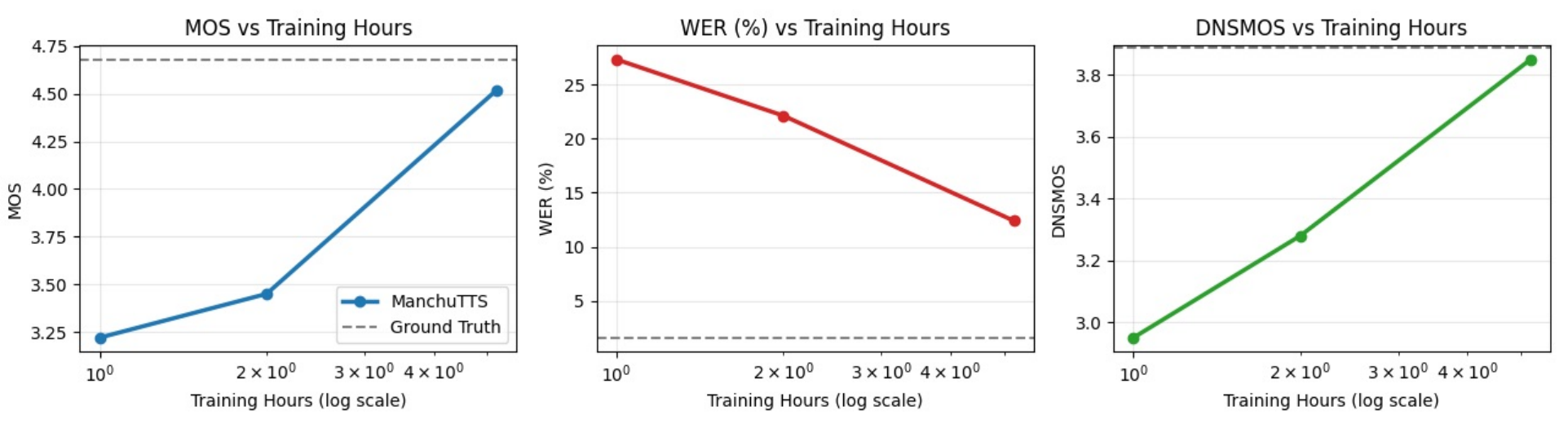}
\caption{Impact of Training Data Size on Manchu Speech Synthesis Performance (Comparison of 1h, 2h, and 5.2h Training Data)}
\label{fig_5}
\vspace{-1.5em}
\end{figure*}

\begin{figure*}[t]
\centering
\includegraphics[width=1\linewidth]{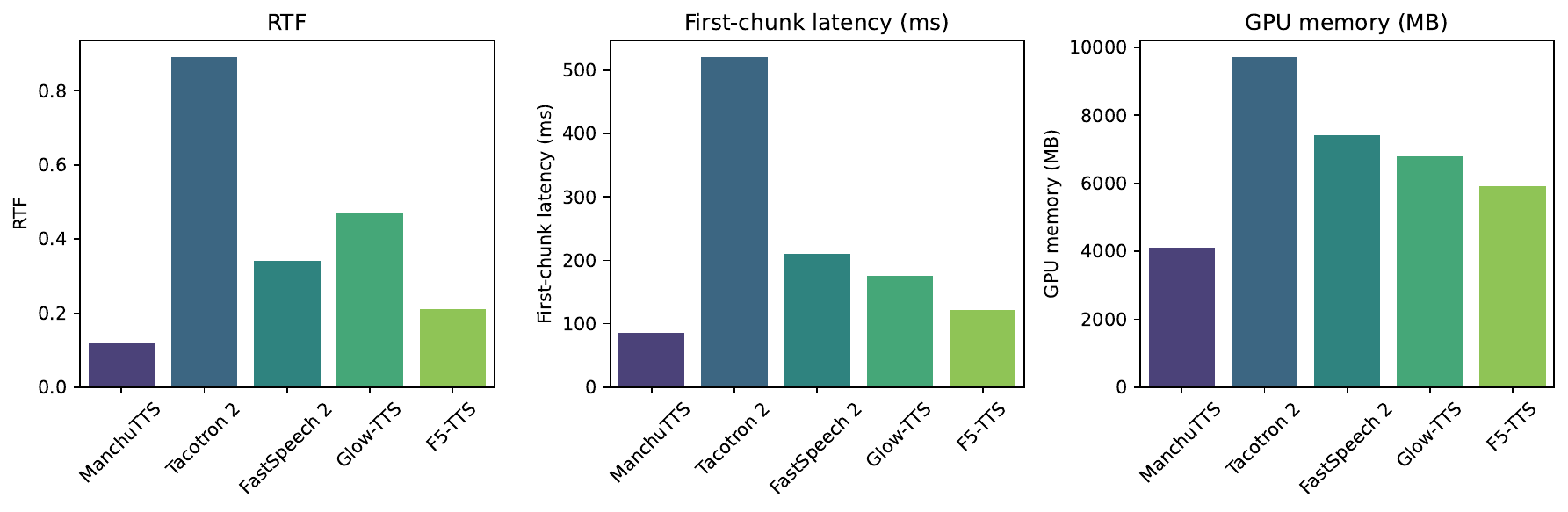}
\vspace{-0.4cm}
\caption{Simulation results for the network.}
\label{fig_7}
\vspace{-0.4cm}
\end{figure*}

\noindent\textbf{Configuration A (Phoneme Layer Only)}: The MOS drops to 3.94, a reduction of 0.58 compared to the full model. Spectral analysis reveals a loss of high-frequency energy ($\geq$4kHz), particularly affecting frication noise in sibilants and nasal resonance in final nasal sounds. Agglutinative word accuracy reaches 70.8\%, and prosodic naturalness is 70.4\%.

\noindent\textbf{Configuration B (Phoneme + Syllable Layers)}: MOS increases to 4.21. Key improvements include smoother stem-suffix transitions, for instance, in the verb "šunggi-ra," the energy gap between the stem /šung/ and suffix /ra/ narrows by 12 ms, contributing to an additional 0.8 dB reduction in MCD. Agglutinative word accuracy rises to 84.7\%, and prosodic naturalness improves to 80.5\%.

\noindent\textbf{Configuration C (Full Three-Layer Model)}: The complete model achieves the highest MOS of 4.52, along with 92.7\% agglutinative word accuracy and 89.4\% prosodic naturalness. Compared to Configuration A, this represents relative improvements of 31\% in word accuracy and 27\% in prosodic naturalness. The prosody layer substantially improves intonation modeling, achieving a 0.91 correlation with native speaker pitch curves in interrogative sentences, compared to 0.63 without it. Additionally, it effectively captures prosodic rules such as "function word weakening," exemplified by the possessive particle "-i," which shortens by 14\% and increases stress by 6\% in sentence-final position.

\subsubsection{Sensitivity Analysis of Data Scale}
We examine how training data size affects synthesis quality using subsets of 1h, 2h, and 5.2h from the full 6.24 hour corpus. As shown in Fig.~\ref{fig_5}, MOS grows nonlinearly from 3.22 (1h) to 4.52 (5.2h), approaching the ground-truth upper bound (4.68). WER drops sharply from 27.3\% to 12.4\% over the same range, while DNSMOS improvements diminish after 2h. Results indicate that about 5 hours represents a practical saturation point for studio-recorded Manchu; further gains would require more diverse or higher-quality field data.

\subsubsection{Cross-Language Transfer and Zero-Shot Generalization}

\begin{figure}[t]
\vspace{-0.2em}
\centering
\includegraphics[width=1\linewidth]{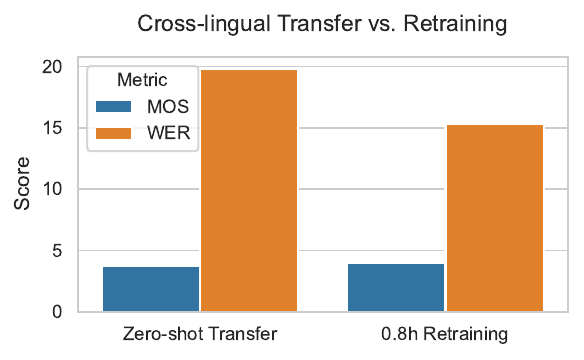}
\caption{Comparison of Ewenki Zero-shot and 0.8 h Retraining Performance}
\label{fig_6}
\vspace{-0.5em}
\end{figure}

ManchuTTS demonstrates effective cross-linguistic transfer in zero-shot Ewenki synthesis (As shown in Table~\ref{tab:5}), achieving a MOS of 3.78±0.15. Which is only marginally below FastSpeech 2 trained directly on 0.8 hours of Ewenki data (4.01±0.13). Despite a higher WER (19.8\% vs 15.3\%). 
As shown in Fig.~\ref{fig_6}, the model preserves key phonetic features, such as vowel formants (r=0.87), and avoids audible degradation common in small-data training, underscoring its practical utility for endangered language support with minimal data.

\subsubsection{Linguistic Verification}
Two Manchu linguists are invited to rate the intelligibility of 50 typical errors (1-5 scale). The results show that samples with missing phonological rules averaged 3.4 points, prosodic boundary misalignment scored 2.8 points, and segment distortion scored 2.1 points, further emphasizing that segment distortion has the greatest impact on intelligibility.

\begin{table}[h]
\vspace{-0.5em}
\centering
\caption{Zero-Shot Results for Ewenki}
\vspace{-0.8em}
\label{tab:5}
\begin{tabular}{|l|c|c|}
\hline
\textbf{Configuration} & \textbf{Manchu $\rightarrow$ Ewenki} & \textbf{Direct Ewenki Training} \\
\hline
MOS & 3.78$\pm$0.15 & 4.01$\pm$0.13 \\
\hline
WER & 19.8$\pm$2.7\% & 15.3$\pm$2.1\% \\
\hline
\end{tabular}
\vspace{-0.5cm}
\end{table}

\subsection{Computational Efficiency and Deployment Analysis}
ManchuTTS demonstrates deployment-ready efficiency, achieving RTF 0.12, 86 ms latency, and 4.1 GB VRAM on an RTX 4090, enabling 8× concurrent streams (Table~\ref{tab:6}). It significantly outperforms autoregressive baselines in speed and resource usage (Fig.~\ref{fig_7}). After INT8 quantization, it sustains 3× real-time synthesis on a Jetson Orin Nano (8 GB), offering a practical edge-computing solution for low-resource language applications.

\subsection{Subjective Listening Tests and Feedback}
The results are shown in Table~\ref{tab:7}. Laboratory tests with 20 native speakers showed ManchuTTS achieved naturalness MOS of 4.52±0.11, comparable to ground truth, and clarity score of 4.61±0.10. Qualitative feedback indicated 63\% elderly listeners praised clear word endings, though limited emotional expression in long sentences and occasional stress drift were noted, highlighting prosodic modeling as key for future improvement.
\begin{table}[t]
\vspace{-0.8em}
\centering
\vspace{0.2em}
\caption{Inference Efficiency (RTX 4090, batch=1, FP16).}
\vspace{-0.8em}
\begin{tabular}{|p{1.5cm}|c|c|c|c|}
\hline
\textbf{Models}
& \textbf{\makecell{RTF\\↓}} 
& \textbf{\makecell{1st-Token\\Lat. (ms)}} 
& \textbf{\makecell{Peak VRAM\\(GB)↓ }} 
& \textbf{\makecell{Concurrency\\↑}} \\ 
\hline
ManchuTTS                & 0.12 & 86  & 4.1 & 8× \\
\hline
\makecell[l]{Tacotron2 \\+WaveGlow}    & 0.89 & 520 & 9.7 & 1× \\
\hline
\makecell[l]{FastSpeech2 \\+HiFi-GAN}    & 0.34 & 210 & 7.4 & 3× \\

\hline
Glow-TTS                 & 0.47 & 175 & 6.8 & 2× \\
\hline
F5-TTS                   & 0.21 & 122 & 5.9 & 5× \\
\hline
\end{tabular}

\label{tab:6}
\vspace{-0.8em} 
\end{table}

\begin{table}[t] 
\vspace{-0.8em}
\centering
\caption{Subjective Listening Test Results (Laboratory Setting, 20 Native Speakers)}
\vspace{-0.8em}
\begin{tabular}{|l|c|c|}
\hline
\textbf{Models} & \textbf{Naturalness} & \textbf{Clarity} \\
\hline
ManchuTTS       & $4.52 \pm 0.11$      & $4.61 \pm 0.10$  \\
\hline
Ground Truth    & $4.68 \pm 0.09$      & $4.53 \pm 0.12$  \\
\hline
Tacotron 2      & $3.21 \pm 0.18$      & $3.40 \pm 0.20$  \\
\hline
FastSpeech 2    & $3.45 \pm 0.15$      & $3.62 \pm 0.17$  \\
\hline
\end{tabular}
\vspace{-0.3em}

\label{tab:7}
\vspace{-0.8em}
\end{table}

\section{Conclusion and Future Work}
ManchuTTS has proven that with 5.2 hours of training data, it can achieve speech synthesis that is acceptable to humans, which has been confirmed by experimental data. However, the model is still limited to studio recorded speech and its accuracy decreases in dialect variations.
Future work will expand data coverage to include different dialects and real-world environments, improve robustness through noise perception training. These steps are crucial for advancing from laboratory validation to actual practical use.

\bibliographystyle{IEEEbib}
\bibliography{icme2026references}
\end{document}